\pdfoutput=1
\documentclass[11pt,a4paper]{article}
\usepackage[hyperref]{naaclhlt2019}
\usepackage{times}
\usepackage{latexsym}

\usepackage{url}

\usepackage{times}
\usepackage{url}
\usepackage{latexsym}
\usepackage{comment}
\usepackage{tikz-dependency}
\usepackage{tikz}
\usepackage{tikz-qtree}
\usepackage[british]{babel}
\usepackage{graphicx}
\usepackage{colortbl}
\usepackage{hhline}
\usepackage{url}
\usepackage{latexsym}
\usepackage{amssymb}
\usepackage{amsmath}
\usepackage{enumerate}
\usepackage{algorithm}
\usepackage{algorithmicx}
\usepackage[noend]{algpseudocode}
\usepackage{booktabs}
\usepackage{proof}
\usepackage{paralist}
\usepackage{tikz-dependency}
\usepackage{tikz} 
\usepackage{graphicx, subfigure}
\usepackage{epstopdf}

\usepackage[latin1]{inputenc}

\usepackage{enumitem}
\setitemize{noitemsep,topsep=0pt,parsep=0pt,partopsep=0pt}
\setenumerate{noitemsep,topsep=0pt,parsep=0pt,partopsep=0pt}

\newcommand{\eat}[1]{}

\makeatletter
\useshorthands{"}%
\defineshorthand{"-}{\nobreak-\bbl@allowhyphens}
\makeatother

\newcommand{\transname}[1]{\ensuremath{\mathsf{#1}}}

\mathchardef\mhyphen="2D 

\newcommand{\sh}{\transname{Shift\mhyphen Attach\mhyphen \textit{p}}}
\newcommand{\re}{\transname{Reduce}}
\newcommand{\at}{\transname{Attach\mhyphen \textit{p}}}

\newcommand{\squishlist}{ 
   \begin{list}{$\bullet$}
    { \setlength{\itemsep}{0pt}      \setlength{\parsep}{3pt} 
      \setlength{\topsep}{3pt}       \setlength{\partopsep}{0pt}
      \setlength{\leftmargin}{1.5em} \setlength{\labelwidth}{1em}
      \setlength{\labelsep}{0.5em} } }

\newcommand{\squishend}{
    \end{list}  }

\aclfinalcopy 
\def\aclpaperid{704} 

\title{Left-to-Right Dependency Parsing with Pointer Networks}

\author{
  Daniel Fern\'{a}ndez-Gonz\'{a}lez\\
  Universidade da Coru\~{n}a\\
  FASTPARSE Lab, LyS Group \\
  Departamento de Computaci\'{o}n \\
  Elvi\~{n}a, 15071 A Coru\~{n}a, Spain \\
  {\tt d.fgonzalez@udc.es} \\
  \\\And
  Carlos G\'{o}mez-Rodr\'{i}guez \\
  Universidade da Coru\~{n}a, CITIC \\
  FASTPARSE Lab, LyS Group \\
  Departamento de Computaci\'{o}n \\
  Elvi\~{n}a, 15071 A Coru\~{n}a, Spain \\
  {\tt carlos.gomez@udc.es}  \\}

\date{}

\begin{document}
\maketitle
\begin{abstract}

We propose a novel transition-based algorithm that straightforwardly parses sentences from left to right by building $n$ attachments, with $n$ being the length of the input sentence. Similarly to the recent stack-pointer parser by \citet{Ma18}, we use the pointer network framework that, given a word, can directly point to a position from the sentence. However, our left-to-right approach is simpler than the original top-down stack-pointer parser (not requiring a stack) and reduces transition sequence length in half, from $2n-1$ actions to $n$. This results in a quadratic non-projective parser that runs twice as fast as the original while achieving the best accuracy to date on the English PTB dataset (96.04\% UAS, 94.43\% LAS) among fully-supervised single-model dependency parsers, and improves over the former top-down transition system in the majority of languages tested.

\end{abstract}

\section{Introduction}
Dependency parsing, the task of automatically obtaining the grammatical structure of a sentence expressed as a dependency tree, has been widely studied by natural language processing (NLP) researchers in the last decades. Most of the models providing competitive accuracies fall into two broad families of approaches: \textit{graph-based} \cite{mcdonald05acl,mcdonald05emnlp}  and \textit{transition-based} \cite{yamada03,Nivre2003} dependency parsers.

Given an input sentence, a graph-based parser scores trees by decomposing them into factors, and performs a search for the highest-scoring tree.

In the past two years, this kind of dependency parsers have 
been ahead in terms of accuracy thanks to
the graph-based neural architecture developed by \citet{DozatM16}, which not only achieved state-of-the-art accuracies on the Stanford Dependencies conversion of the English Penn Treebank (hereinafter, PTB-SD), but also
obtained the best results in the majority of languages
in the CoNLL 2017 Shared Task \cite{DozatQM17}. This tendency recently changed, since a transition-based parser developed by \citet{Ma18} managed to outperform the best graph-based model in 
the majority of
datasets tested. 

Transition-based parsers incrementally build a dependency graph for an input sentence by 
applying a sequence of 
transitions. 
This results in more efficient parsers with linear time complexity for parsing projective sentences, or quadratic for handling non-projective structures, when implemented with greedy or beam search. However, their main weakness is 
the lack of access to global context information when transitions are greedily chosen. This 
favours error propagation, mainly affecting long dependencies that require a larger number of transitions to be built \cite{McDonald2011analyzing}. 

Many attempts have been made to alleviate the impact of error propagation in transition-based dependency parsing, but the latest and most successful approach was developed by \citet{Ma18}. In particular, they make use of \textit{pointer networks} \cite{Vinyals15} to implement a new neural network architecture called \textit{stack-pointer network}.
The proposed framework provides a global view of the input sentence by capturing information from the whole sentence and all the arcs previously built, crucial for reducing the effect of error propagation; and, thanks to an attention mechanism \cite{Bahdanau2014,Luong2015}, is able to return a position in that sentence that corresponds to a word related to the word currently on top of the stack. They take advantage of this and propose a novel transition system that 
follows a top-down depth-first strategy to perform the syntactic analysis. Concretely, it considers the word pointed by the neural network as the child of the word on top of the stack, 
and builds the corresponding dependency relation between them.
This results in a transition-based algorithm that can process unrestricted non-projective sentences in 
$O(n^2)$
time complexity and requires 2$n$-1 actions to successfully parse a sentence with $n$ words. 

We also take advantage of pointer network capabilities and use the neural network architecture introduced by \citet{Ma18} to design a non-projective left-to-right transition-based algorithm, where the position value pointed by the network has the opposite meaning: it denotes the index that corresponds to the head node of the current focus word.  
This results in a straightforward transition system that can parse a sentence in just $n$ actions, without the need of any additional data structure and by just attaching each word from the sentence to another word (including the root node). Apart from increasing the parsing speed twofold (while keeping the same quadratic time complexity), it achieves the best accuracy to date among fully-supervised single-model dependency parsers on the PTB-SD, and obtains competitive accuracies on twelve different languages in comparison to the original top-down version. 

\section{Preliminaries}
\citet{Ma18} propose a novel neural network architecture whose main backbone is a pointer network \cite{Vinyals15}. This kind of neural networks are able to learn the conditional probability of a sequence of discrete numbers that correspond to positions in an input sequence (in this case, indexes of words in a sentence) and, by means of attention \cite{Bahdanau2014, Luong2015}, implement a pointer that selects a position from the input at decoding time.

Their approach initially reads the whole sentence, composed of the $n$ words $w_1, \dots ,w_n$, and encodes each $w_i$ 
one by one into an \textit{encoder hidden state} $e_i$.
As encoder, they employ a combination of CNNs and bi-directional LSTMs \cite{Chiu2016, Ma2016}. For each word, CNNs are used to obtain its character-level representation that is concatenated to the word and PoS embeddings to finally be fed into BiLSTMs that encode word context information.

As decoder they present a top-down transition system, where parsing configurations 
use the classic data structures \cite{nivre08cl}: a \textit{buffer} (that contains unattached words) and a \textit{stack} (that holds partially processed words). 

The available parser actions are
two transitions that we call $\sh$ and $\re$. Given a configuration with word $w_i$ on top of the stack, as the pointer network just returns a position $p$ from a given sentence, they proceed as follows to determine which transition should be applied: 
\begin{itemize}
    \item If $p \neq i$,
    then the pointed word $w_p$ is considered as a child of $w_i$; so the parser chooses a $\sh$ transition to move $w_p$ from the buffer to the stack and build an arc $w_i \rightarrow w_p$.
    \item On the other hand,  
    if $p = i$,
    then 
    $w_i$ is considered to have found all its children, and
    a $\re$ transition is applied to pop the stack.
\end{itemize}

\noindent The parsing process starts with a dummy root \$ on 
the stack and, by applying 2$n$-1 transitions, a dependency tree is built for the input 
in a top-down depth-first fashion, where multiple children of a same word are forced during training to be created in an inside-out manner. More in detail, for each parsing configuration $c_t$, the decoder (implemented as a uni-directional LSTM) receives the encoder hidden state $e_i$ of the word $w_i$ on top of the stack to generate a \textit{decoder hidden state} $d_t$. After that, $d_t$, together with the sequence $s_i$ of encoder hidden states from words still in the buffer plus $e_i$, are used to compute the attention vector $a^t$ as follows: 
\begin{align}
v^t_i = score(d_t, s_i)\\
a^t = softmax(v^t)
\end{align}
As attention scoring function ($score()$), they adopt the biaffine attention mechanism described in \cite{Luong2015, DozatM16}. Finally, the attention vector $a^t$ will be used to return the highest-scoring position $p$
and choose the next transition. 
The parsing process ends when 
only the root remains on the stack.

As extra high-order features, \citet{Ma18} add grandparent and sibling information, 
whose encoder hidden states are added to that of the word on top of the stack to generate the corresponding decoder hidden state $d_t$.
They prove that these additions improve final accuracy, especially when children are attached in an inside-out fashion.

According to the authors, the original stack-pointer network is trained to maximize the likelihood of choosing the correct word for each possible top-down path from the root to 
a leaf.
More in detail, a dependency tree can be represented as a sequence of top-down paths $p_1, \dots , p_k$, where each path $p_i$ corresponds to a sequence of words $\$, w_{i,1}, w_{i,2}, \dots, w_{i,l_i}$ from the root to a leaf. Thus, the conditional probability $P_\theta (y|x)$ of the dependency tree $y$ for an input sentence $x$ can be factorized according to this top-down structure as:
\begin{align*}
P_\theta (y|x) &= \prod_{i=1}^k P_\theta (p_i | p_{<i}, x) \\
&= \prod_{i=1}^k \prod_{j=1}^{l_i} P_\theta (w_{i,j} | w_{i,<j},p_{<i},x) 
\end{align*}

\noindent where $\theta$ represents model parameters, $p_{<i}$ stands for previous paths already explored, $w_{i,j}$ denotes the $j$th word in path $p_i$ and $w_{i,<j}$ represents all the previous words on $p_i$. 

For more thorough details of the stack-pointer network architecture and the top-down transition system, please read the original work by \citet{Ma18}.

\section{Our approach}
We take advantage of the 
neural network architecture designed by \citet{Ma18} and introduce a simpler left-to-right transition system that requires neither a stack nor a buffer to process the input sentence and where, instead of selecting a child of the word on top of the stack, the network points to the parent of the current focus word.

In particular, in our proposed approach, the parsing configuration just corresponds to a \textit{focus word pointer} $i$, that is used to point to the word currently being processed. The decoding process starts with $i$ pointing at the first word of the sentence and, at each parsing configuration,
only one action is available: the parameterized $\at$ transition, that 
links the focus word $w_i$ to the head word $w_p$ in position $p$ of the sentence (producing the dependency arc $w_p \rightarrow w_i$) and moves $i$ one position to the right. Note that, 
in our algorithm, $p$ can equal 0, attaching, in that case, $w_i$ to the dummy root node. The parsing process ends when the last word from the sentence is attached.
This can be easily represented as a loop that traverses the input sentence from left to right, linking each word to another from the same sentence or to the dummy root. Therefore, we just need $n$ steps to process the $n$ words of a given sentence and build a dependency tree.

While our novel transition system intrinsically holds the single-head constraint (since, after attaching the word $w_i$, $i$ points to the next word $w_{i+1}$ in the sentence), it can produce an output with cycles.\footnote{In practice, even with the cycle detection mechanism disabled, the presence of cycles in output parses is very uncommon (for instance, just in 1\% of sentences in the PTB-SD dev set) since our system seems to adequately model well-formed tree structures.} Therefore, in order to build a well-formed dependency tree during decoding, attachments that generate cycles in the already-built dependency graph must be forbidden. Please note that the need of a cycle-checking extension does not increase the overall quadratic runtime complexity of the original implementation by \citet{Ma18} since, as in other transition-based parsers such as \cite{covington01fundamental,gomniv2010}, cycles can be incrementally identified in amortized constant time by keeping track of connected components using path compression and union by rank. Therefore, the left-to-right algorithm requires $n$ steps to produce a parse. In addition, at each step, the attention vector $a_t$ needs to be computed and cycles must be checked, both in $O(n)+O(n)=O(n)$ runtime. This results in a $O(n^2)$ time complexity for decoding.\footnote{A practically faster version of the left-to-right parser might be implemented by just ignoring the presence of cycles during decoding, and destroying the cycles generated as a post-processing step that simply removes one of the arcs involved.} 

 On the other hand, while in the top-down decoding only available words in the buffer (plus the word on top of the stack) can be pointed to by the network and they are reduced as arcs are 
 created (basically to keep the single-head constraint); our proposed approach is less rigid: all words from the sentence (including the root node and excluding $w_i$) can be pointed to, 
 as long as they satisfy the acyclicity constraint. This is necessary because two different words might be attached to the same head node and the latter can be located in the sentence either before or after $w_i$. Therefore, the sequence $s_i$, required by the attention score function (Eq.(1)), is composed of the encoder hidden states of all words from the input, excluding $e_i$, and prepending a special vector representation denoting the root node.

We also add extra features to represent the current focus word. Instead of using grandparent and sibling information (more beneficial for a top-down approach), we just add the encoder hidden states of the previous and next words in the sentence to generate $d_t$, which seems to be more suitable for a left-to-right decoding.

In dependency parsing, a tree for an input sentence of length $n$ can be represented
as a set of $n$ directed and binary links $l_1, \dots , l_{n}$. Each link $l_i$ is characterized by the word $w_i$ in position $i$ in the sentence and its head word $w_h$, resulting in a pair $(w_i, w_h)$. Therefore, to train this novel variant, we factorize the conditional probability $P_\theta (y|x)$ to a set of head-dependent pairs as follows:
\begin{align*}
P_\theta (y|x) &= \prod_{i=1}^{n} P_\theta (l_i | l_{<i}, x) \\
&= \prod_{i=1}^{n} P_\theta (w_h | w_i,l_{<i},x) \end{align*}

\noindent Therefore, the left-to-right parser is trained by maximizing the likelihood of choosing the correct head word $w_h$ for the word $w_i$ in position $i$, given the previous predicted links $l_{<i}$.

Finally, following a widely-used approach (also implemented in \cite{Ma18}), dependency labels are predicted by a multiclass classifier, which is trained in parallel with the parser by optimizing the sum of their objectives.

\section{Experiments}
\subsection{Data and Settings}
We use the same implementation as \citet{Ma18} and conduct experiments on the Stanford Dependencies \cite{deMarneffe2008} conversion (using the Stanford parser v3.3.0)\footnote{\url{https://nlp.stanford.edu/software/lex-parser.shtml}} of the English Penn Treebank \cite{marcus93}, with standard splits and predicted PoS tags. In addition, we compare our approach to the original top-down parser on the same twelve languages from the Universal Dependency Treebanks\footnote{\url{http://universaldependencies.org}} (UD) that were used by \citet{Ma18}.\footnote{Please note that, since they used a former version of UD datasets, we reran also the top-down algorithm on the latest treebank version (2.2) in order to perform a fair comparison.} 

Following standard practice, we just exclude punctuation for evaluating on PTB-SD and, for each experiment, we report the average Labelled and Unlabelled Attachment Scores (LAS and UAS) over 3 and 5 repetitions for UD and PTB-SD, respectively. 

Finally, we use the same hyper-parameter values, pre-trained word embeddings and beam size (10 for PTB-SD and 5 for UD) as \citet{Ma18}.

\begin{table}
\begin{small}
\begin{center}
\centering
\begin{tabular}{@{\hskip 0pt}lll@{\hskip 0pt}}
Parser & UAS & LAS \\
\hline
\citet{CheMan2014} &   91.8  &  89.6  \\
\citet{Dyer2015} &  93.1 & 90.9  \\
\citet{Weiss2015} & 93.99  &  92.05  \\
\citet{Ballesteros2016}  &  93.56  &  91.42  \\
\citet{Kiperwasser2016} & 93.9 & 91.9 \\
\citet{Alberti2015}  & 94.23  &  92.36  \\
\citet{Qi2017} & 94.3 & 92.2 \\
\defcitealias{Fernandez18}{Fern\'andez-G and G\'omez-R (2018)}\citetalias{Fernandez18}
& 94.5 &  92.4  \\
\citet{Andor2016}  & 94.61  &  92.79  \\
\citet{Ma18}$^*$ & 95.87 & 94.19 \\
\textbf{This work}$^*$ &  \textbf{96.04} &  \textbf{94.43}  \\
\hline
\citet{Kiperwasser2016} & 93.1 & 91.0 \\
\citet{Wang2016} & 94.08 & 91.82 \\
\citet{Cheng2016} & 94.10 & 91.49 \\
\citet{Kuncoro2016} & 94.26 & 92.06 \\
\citet{Zhang17} & 94.30 & 91.95 \\
\citet{Ma2017} & 94.88 & 92.96 \\
\citet{DozatM16} & 95.74 & 94.08 \\ 
\citet{Ma18}$^*$ & 95.84 & 94.21 \\
\hline
\multicolumn{1}{c}{}\\
\end{tabular}
\centering
\setlength{\abovecaptionskip}{4pt}
\caption{Accuracy comparison of state-of-the-art fully-supervised single-model dependency parsers on PT-SD. The first block contains transition-based algorithms and the second one, graph-based models. Systems marked with $^*$, including the improved variant described in \cite{Ma18} of the graph-based parser by \cite{DozatM16}, are implemented under the same framework as our approach and use the same training settings. Like \cite{Ma18}, we report the average accuracy over 5 repetitions.}
\label{tab:ptb}
\end{center}
\end{small}
\end{table}

\subsection{Results}
By outperforming the two current state-of-the-art graph-based \cite{DozatM16} and transition-based \cite{Ma18} models on the PTB-SD, our approach becomes the most accurate
fully-supervised
dependency parser developed so far, as shown in Table~\ref{tab:ptb}.\footnote{It is worth mentioning that all parsers reported in this section make use of pre-trained word embeddings previously learnt from corpora beyond the training dataset. However, it is common practice in the literature that systems that only use standard pre-trained word embeddings are classed as fully-supervised models, even though, strictly, they are not trained exclusively on the official training data.}

\begin{table}
\begin{small}
\begin{center}
\centering
\begin{tabular}{@{\hskip 0pt}l@{\hskip 0.3pt}|@{\hskip 2pt}cc@{\hskip 2pt}|@{\hskip 2pt}cc@{\hskip 0.3pt}}
& \multicolumn{2}{@{\hskip 0.3pt}c|@{\hskip 2pt}}{Top-down}
& \multicolumn{2}{@{\hskip 0.8pt}c@{\hskip 0.3pt}}{Left-to-right}
\\
& UAS & LAS
& UAS & LAS 
\\
\hline
bu & \textbf{94.42$\pm$0.02} & \textbf{90.70$\pm$0.04} & 94.28$\pm$0.06  & 90.66$\pm$0.11 \\ 
ca & 93.83$\pm$0.02 & 91.96$\pm$0.01 & \textbf{94.07$\pm$0.06} & \textbf{92.26$\pm$0.05} \\ 
cs & 93.97$\pm$0.02 & 91.23$\pm$0.03 & \textbf{94.19$\pm$0.04}  & \textbf{91.45$\pm$0.05} \\
de & \textbf{87.28$\pm$0.07} & \textbf{82.99$\pm$0.07} & 87.06$\pm$0.05 & 82.63$\pm$0.01  \\ 
en & 90.86$\pm$0.15 & 88.92$\pm$0.19  & \textbf{90.93$\pm$0.11}  &  \textbf{88.99$\pm$0.11} \\ 
es & 93.09$\pm$0.05 & 91.11$\pm$0.03 & \textbf{93.23$\pm$0.03} & \textbf{91.28$\pm$0.02} \\
fr & \textbf{90.97$\pm$0.09} & \textbf{88.22$\pm$0.12}  & 90.90$\pm$0.04 & 88.14$\pm$0.10  \\ 
it & 94.08$\pm$0.04 & 92.24$\pm$0.06 & \textbf{94.28$\pm$0.06} & \textbf{92.48$\pm$0.02}   \\
nl & \textbf{93.23$\pm$0.09} & 90.67$\pm$0.07  & 93.13$\pm$0.07 & \textbf{90.74$\pm$0.08}  \\ 
no & 95.02$\pm$0.05 & 93.75$\pm$0.05 & \textbf{95.23$\pm$0.06} & \textbf{93.99$\pm$0.07} \\ 
ro & 91.44$\pm$0.11 & 85.80$\pm$0.14 & \textbf{91.58$\pm$0.08} & \textbf{86.00$\pm$0.07} \\ 
ru & 94.43$\pm$0.01 & 93.08$\pm$0.03 & \textbf{94.71$\pm$0.07} & \textbf{93.38$\pm$0.09} \\ 
\hline
\multicolumn{5}{c}{}\\
\end{tabular}
\centering
\setlength{\abovecaptionskip}{4pt}
\caption{Parsing accuracy of the top-down and left-to-right pointer-network-based parsers on test datasets of twelve languages from UD. Best results for each language are shown in bold and, apart from the average UAS and LAS, we also report the corresponding standard deviation over 3 runs.}
\label{tab:ud}
\end{center}
\end{small}
\end{table}

In addition, in Table~\ref{tab:ud} we can see how, under the exactly same conditions, the left-to-right algorithm improves over the original top-down variant in nine out of twelve languages in terms of LAS, obtaining competitive results in the remaining three datasets. 

Finally, in spite of requiring a cycle-checking procedure, our approach proves to be twice as fast as the top-down alternative in decoding time, achieving, under the exact same conditions, a 23.08-sentences-per-second speed on the PTB-SD compared to 10.24 of the original system.\footnote{Please note that the implementation by \citet{Ma18}, also used by our novel approach, was not optimized 
for speed
and, therefore, the reported speeds are just intended for comparing algorithms implemented under the same framework, but not to be considered as the best speed that a pointer-network-based system can potentially achieve.}

\section{Related work}
There 
is previous work that proposes to implement dependency parsing by independently selecting the head of each word in a sentence, using neural networks.
In particular, \citet{Zhang17} make use of a BiLSTM-based neural architecture to 
compute the probability of attaching each word to one of the other input words, in a similar way as pointer networks do. 
During decoding, a post-processing step is needed to produce well-formed trees by means of a maximum spanning tree algorithm. 
Our approach does not need this post-processing, as cycles are forbidden during parsing instead, and achieves a higher accuracy thanks to the pointer network architecture and the use of information about previous dependencies.

Before \citet{Ma18} presented their top-down parser, \citet{Chorowski17} had already employed pointer networks \cite{Vinyals15} for dependency parsing. Concretely, they developed a pointer-network-based neural architecture with multitask learning able to perform pre-processing, tagging and dependency parsing exclusively by reading tokens from an input sentence, 
without needing
POS tags or pre-trained word embeddings. Like our approach, they also use the capabilities provided by pointer networks to undertake the parsing task as a simple process of attaching each word 
as dependent of another.
They also try to improve the network performance with POS tag prediction as auxiliary task and with different approaches to perform label prediction. 
They do not exclude cycles, neither by forbidding them at parsing time or by removing them by post-processing, as they report that their system produces parses with a negligible amount of cycles, even with greedy decoding (matching our observation for our own system, in our case with beam-search decoding).
Finally, the system developed by \citet{Chorowski17} is constrained to projective dependencies, while our approach can handle unrestricted non-projective structures.

\section{Conclusion}
We present a novel left-to-right dependency parser based on pointer networks. We follow the same neural network architecture as the stack-pointer-based approach developed by \citet{Ma18}, but just using a focus word index instead of a buffer and a stack. Apart from doubling their system's speed, our approach proves to be a competitive alternative on a variety of languages and achieves the best accuracy to date on the PTB-SD. 

The good performance of our algorithm can be explained by the shortening of the transition sequence length. In fact, it has been proved by several studies \cite{buffertrans, Qi2017, Fernandez18} that by reducing the number of applied transitions, the impact of error propagation is alleviated, yielding more accurate parsers.

Our system's source code is freely available at \url{https://github.com/danifg/Left2Right-Pointer-Parser}.

\section*{Acknowledgments}

This work has received funding from the European
Research Council (ERC), under the European
Union's Horizon 2020 research and innovation
programme (FASTPARSE, grant agreement No
714150), from 
MINECO (FFI2014-51978-C2-2-R, TIN2017-85160-C2-1-R)
and from Xunta de Galicia (ED431B 2017/01).

\bibliography{main,twoplanaracl,bibliography}
\bibliographystyle{acl_natbib}

\end{document}